\documentclass[a4paper,twocolumn]{article} 
\usepackage[margin=25.4truemm]{geometry}
\usepackage{cite}
\pdfoutput=1
\usepackage{caption}
\usepackage[pdftex]{graphicx}
\usepackage{indentfirst}
\usepackage{breqn}
\usepackage{textcomp}
\usepackage{multirow}
\usepackage{amsmath,amssymb}
\usepackage{bbm}
\newcommand{\ctext}[1]{\raise0.2ex\hbox{\textcircled{\scriptsize{#1}}}}

\makeatletter

\def\JTeX{\leavevmode\lower .5ex\hbox{J}\kern-.17em\TeX}
\def\JLaTeX{\leavevmode\lower.5ex\hbox{J}\kern-.17em\LaTeX}
\makeatother
\usepackage{color, hyperref}
\hypersetup{
	setpagesize=false,
	bookmarksnumbered=true,
	bookmarksopen=true,
	colorlinks=true,
	linkcolor=red,
	citecolor=green,
}
\usepackage{titlesec}
\titleformat*{\section}{\Large\bfseries\sffamily}
\titleformat*{\subsection}{\large\bfseries\sffamily}
\titleformat*{\subsubsection}{\large\bfseries\sffamily}

\renewenvironment{abstract}
{\quotation\small\noindent\rule{\linewidth}{.5pt}\par\smallskip
	{\centering\bfseries\abstractname\par}\medskip}
{\par\noindent\rule{\linewidth}{.5pt}\endquotation}
\providecommand{\keywords}[1]
{\textbf{\textit{Key words: }} #1}

\begin{document}
	\setlength{\baselineskip}{10.5pt}
	\setlength{\leftskip}{2truemm}
	\setlength{\rightskip}{2truemm}
	\twocolumn[
	\begin{center}
		\vspace{0em}
		{\LARGE\bfseries 3D-Plotting Algorithm for Insects using YOLOv5\par\vspace{\baselineskip}}
		Daisuke Mori\textsuperscript{1}, 
		Hiroki Hayami\textsuperscript{2},
		Yasufumi Fujimoto\textsuperscript{2}
		and
		Isao Goto\textsuperscript{3}
		\par\vspace{\baselineskip}
		\textsuperscript{1} Miyagi University Graduate School of Food, Agricultural and Environmental Sciences, Japan\par 
		\textsuperscript{2} The Miyagi Prefectural Izunuma-Uchinuma Environmental Foundation, Japan\par
		\textsuperscript{3} Miyagi University School of Food Industrial Sciences, Japan
	\end{center}
	\begin{abstract}
		\noindent
		\setlength{\baselineskip}{10.5pt}
		In ecological research, accurately collecting spatiotemporal position data is a fundamental task for understanding the behavior and ecology of insects and other organisms. In recent years, advancements in computer vision techniques have reached a stage of maturity where they can support, and in some cases, replace manual observation. In this study, a simple and inexpensive method for monitoring insects in three dimensions (3D) was developed so that their behavior could be observed automatically in experimental environments. The main achievements of this study have been to create a 3D monitoring algorithm using inexpensive cameras and other equipment to design an adjusting algorithm for depth error, and to validate how our plotting algorithm is quantitatively precise, all of which had not been realized in conventional studies. By offering detailed 3D visualizations of insects, the plotting algorithm aids researchers in more effectively comprehending how insects interact within their environments.
	\end{abstract}
	\newline
	]
	
	\section{Introduction}
		In ecological research, accurately collecting spatiotemporal position data is a fundamental task for understanding the behavior and ecology of insects and other organisms. Traditionally, this tracking has relied primarily on manual observation and recording of positional information. For example, studies in animal behavior and ecology often involved tagging animals with unique identifiers and manually observing them~\cite{Thomas1991}. These datasets have been invaluable for advancing these fields, but manual data collection is time-consuming and labor-intensive. Moreover, the weight of tags may affect the behavior of small or lightweight insects~\cite{Kissling2014}, limiting their implementation. \par
		To improve these problems, several researchers have proposed a wide variety of automated trackers for insects or small animals to make an invaluable tool for collecting large amounts of activity data in a hands-free, cost-effective, and non-invasive manner. Crall et al. (2015)~\cite{Crall2015} developed tracking visual tags named BEEtag that makes identification through separate images or movie frames at low-cost. This method makes for a tracking system that not only avoids error propagation but also performs well in images with complex backgrounds. However, utilizing tags~\cite{Dennis2008} can significantly affect stress levels of the insects~\cite{Sockman2001} and their behavior~\cite{Coudert2005}. Sondhi et al. (2021)~\cite{Soudhi2022} developed a portable locomotion activity monitor (pLAM), or a mobile activity detector, to quantify small insect activity; pLAM automatically detects and counts insect’s activity events with infrared lights, inexpensive components, and open-source motion tracking software, but it cannot run effectively when there is interference such as background motion which can negatively affect the results. This is a major drawback when used in noisy environments.\par
		In recent years, the latest developments in deep learning, computer vision, and image processing have evolved enough to assist, and sometimes even take the place of, manual observation. Bjerge et al. (2022)~\cite{Bjerge2022} developed a system for detecting, tracking, and identifying individual insects in situ constructed from commercial off-the-shelf components and used deep learning to perform species detection and classification from images. \par
		Recently, insect monitoring systems in three dimensions (3D) are being developed to track insect movements more comprehensively. Choi et al. (2023)~\cite{Choi2023} made a model system with two cameras that tracks insects in 3D to identify them and record their behavior, movement, size, and habits with image processing and deep learning. However, issues such as the effect of depth error on accuracy were not considered in the study. Moreover, the precision of the system in tracking insects in virtual 3D space was not tested. Few studies have coped with such limitations in precisely observing insect behavior in the laboratory and have not been able to validate monitored results. \par
		The main purpose of our research was to develop a simple and inexpensive method for monitoring insects in 3D to observe their behavior automatically in experimental environments. The key points of our study are summarized as follows: 
	\begin{itemize}
		\item{We developed a 3D insect plotting algorithm with deep learning detector and inexpensive components. }
		\item{We proposed an adjusting algorithm to alleviate depth error which occurs inevitably on 3D monitoring. This makes 3D plotting of insects more precise. }
		\item{We validated the precision of the system for plotting insects in virtual 3D space quantitatively in experimental environments.}
	\end{itemize}
	\section{Materials and Methods}
	\begin{figure}[htbp]
		\centering
		\includegraphics[width=1.0\columnwidth]{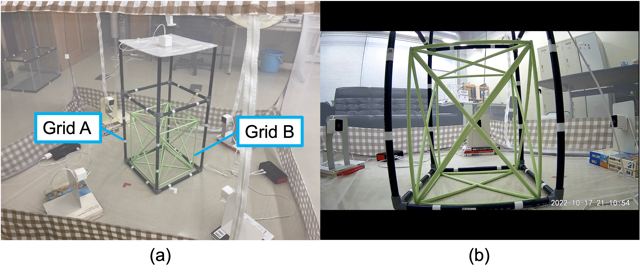}
		\vspace{-1.8em}
		\caption{ Set up of the observation system (a) and a frame from the recorded videos (b). White stickers attached to the black framework (Grid A) are marks for algorithm calculation.}
		\label{fig:setup}
		\vspace{0em}
	\end{figure}
	Our observation environment is shown in Figure~\ref{fig:setup}(a). A smaller three-dimensional green grid made of paper called Grid B is set inside a larger three-dimensional black grid consisting of polyvinyl chloride pipes (PVC pipes) named Grid A. Insects are set on Grid B, which is used in testing the precision of our plotting algorithm, while Grid A has white markings that serve as placement markers used in identifying and plotting the insects’ locations. Additionally, the setup involves five precisely aligned monocular cameras with video recording devices. These cameras are strategically positioned to observe consecutive frames of insect movements, capturing and storing this data into a dedicated storage system. In each frame, insect detectors are used to determine the precise positions of the insects. Based on the system setup details, including cameras and insect position within the frame, it becomes feasible to estimate the 3D location of the insects in world coordinates system for that specific frame. The 3D location is stored in a CSV file. Then, a 3D reconstruction algorithm is run to visualize insects in virtual 3D space. The programs for 3D plotting were all written in Python (version 3.8.10).
	\subsection{Hardware Details}
		Grid A is not only used for monitoring movements of the insects but also needed to estimate the insects’ 3D location. Grid A’s dimensions in millimeters are 390W $\times$ 850H $\times$ 390D. It has twenty white marks (Figure~\ref{fig:setup}(b)) which are used in adjusting and unifying the cameras’ positions and in modifying plotting error due to depth effect.\par
		The five monocular cameras within this system are employed to construct a 3D representation of the insects’ positions and movements. The four of five cameras are placed at about 245 mm height from the ground equidistant from each side of the Grid A. These four cameras are referred to as the side cameras. Although the rule of equidistance should be followed, it is not necessary to be overly sensitive about the rule because error occurring from the arrangement of cameras is modified using our adjusting algorithm, which is described in Section~\ref{3DPlottingAlgorithm}. Each of the four side cameras is attached with a steel bookstand using magnetic force. The remaining camera is placed at the top of Grid A to record the insects’ movements and some of the white markers. This fifth camera is referred to as the top camera. We chose ATOM Cam 2~\cite{ATOM2023} as a camera utilized for this research. It is an inexpensive internet protocol monocular camera which records 20 frames per second (fps) on bright mode and 15 fps on dark mode using an infrared system. Recorded movies are saved on a micro-SD memory with timestamp. We selected a high-resolution camera (1920 $\times$ 1080 pixels) to capture clear videos of the insects due to their small size. These cameras offer a wide-angle view of 120\textdegree, facilitating comprehensive coverage. 
	\subsection{Insect Detection using Deep Learning}
		Convolutional neural networks (CNNs) are widely employed for object detection~\cite{Indolia2018,Shrestha2019,Liu2020,Zhao2019} across various domains, including applications such as insects detection and species identification. CNNs predict bounding boxes around objects in the image along with their corresponding class labels and confidence scores. Object detectors using CNNs are classified into two types: two-stage detectors which infer classification of the object after detection, and one-stage detectors which make inference of detection and classification at the same time. ”You Only Look Once” (YOLO)~\cite{Redmon2015} model which stands out as a one-stage object detector is widely used across various applications, including utilization in insects detection~\cite{Bjerge2022}. Two-stage detectors like the Faster Region-based Convolutional Neural Network (Faster R-CNN)~\cite{Ren2017} are also widely used and have been adopted specifically to detect small objects~\cite{Cao2019}.\par
		While there are lots of detectors using CNNs, we chose YOLOv5~\cite{Jocher2021} because it is one of the most reliable detectors for insects~\cite{Teixeira2023}. In terms of both fps and mean average precision (mAP), YOLOv5 has improved performance compared to the previous versions. Unlike other YOLO models, YOLOv5 has various models which are divided by size depending on depth and layer width. Balancing accuracy and speed are often challenging as they are typically incompatible. YOLOv5s, the fastest model, sacrifices its inference accuracy, while the YOLOv5x model operates more slowly but achieves higher accuracy. In this research, YOLOv5s model was applied because it does not only train faster than other models but also achieves enough detection accuracy to estimate the insect’s 3D location. The architecture of YOLOv5 consists of three primary components: the backbone, neck, and head. The backbone is the main body of the network for feature extraction. YOLOv5 uses CSPNet-based CSP-Darknet53 as the backbone. The neck represents the intermediary layers situated between the backbone and the head. Typically, these layers are utilized to gather feature maps from various stages. In YOLOv5, SPPF and New CSP-PAN structures are utilized. The head has the purpose of generating the final output. Initially, this model sets the anchor box, utilizing it as a base to generate the bounding box. Like YOLOv3~\cite{Joseph2018}, it generates bounding boxes across three scales, employing three anchor boxes for each scale. Although we decided on the YOLOv5 as the insect detector of choice, it is worth noting that many kinds of reliable detectors for insects such as Faster R-CNN could also be adopted instead of YOLOv5 because all we need to estimate insect 3D coordinates is to get the 2D coordinates from images. 
	\subsubsection{Dataset Collection and Labeling}
\begin{figure}[htbp]
			\centering
			\includegraphics[width=0.9\columnwidth]{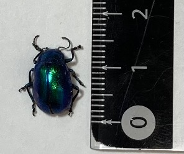}
			\vspace{-0.5em}
			\caption{\textit{Chrysolina virgata.}}
			\label{fig:insect}
			\vspace{-1em}
		\end{figure}

		To build a dataset for training YOLOv5, we recorded videos of the insects’ activity using our original equipment (Figure~\ref{fig:setup}(a)). We chose the leaf insect \textit{Chrysolina virgata} (Figure~\ref{fig:insect}) as the experimental insect because the data acquisition is easy due to their lack of ability to fly and move long distances. \textit{Chrysolina virgata} is listed as an endangered species in Japan. The insects, captured from Lake Izunuma-Uchinuma, where there is a population of \textit{Chrysolina virgata}, are protected by local environmental groups, so they were not harmed. At the end of the experiment, the insects were returned to the site where they were captured.\par
		Using the five cameras simultaneously, we recorded 34.7 minutes of a single insect; an additional 12 minutes of recording was then taken after adding a second insect. The single-insect videos were then separated into different segments, with one segment per location where the insect moved around. In the two-insect video, a single frame was selected from every 0.5 second (or every 10 frames) of recording. After deleting the frames which contained no images of insects, we obtained 4029 frames or images from the two-insect video with a total of 5400 pictured insects from the five cameras combined. We divided this dataset into 4852 and 548 pictured insects, corresponding to the training and testing sets, respectively. Data Augmentation (DA) including flip, rotation, and Gaussian noise (mean = 0, variance = 3) was applied to the training set to enhance robustness of detections. We set all DA probabilities at 0.5 to reflect how often it was applied. After processing DA, background-only images that contained no insects, which numbered 5\% of whole training set, were added to help reduce False Positive (FP) following the YOLOv5 manual~\cite{Jocher2021}. Finally, we obtained a dataset consisting of 6868, 764, and 403 images (8728, 976, and 548 insects) respectively for training, validation, and testing.\par
		We used a labeling tool (Image annotation tool VoTT ver 1.7.2~\cite{VoTT2023}) to annotate the dataset. For every insect in each image, we made a bounding box which was specifically designed to include the entire insect while excluding any unnecessary background elements.\par
	\subsubsection{Training}
	\begin{figure*}[htbp]
		\begin{minipage}{\linewidth}
			\centering
			\vspace{-1.5em}
			\includegraphics[width=0.85\columnwidth]{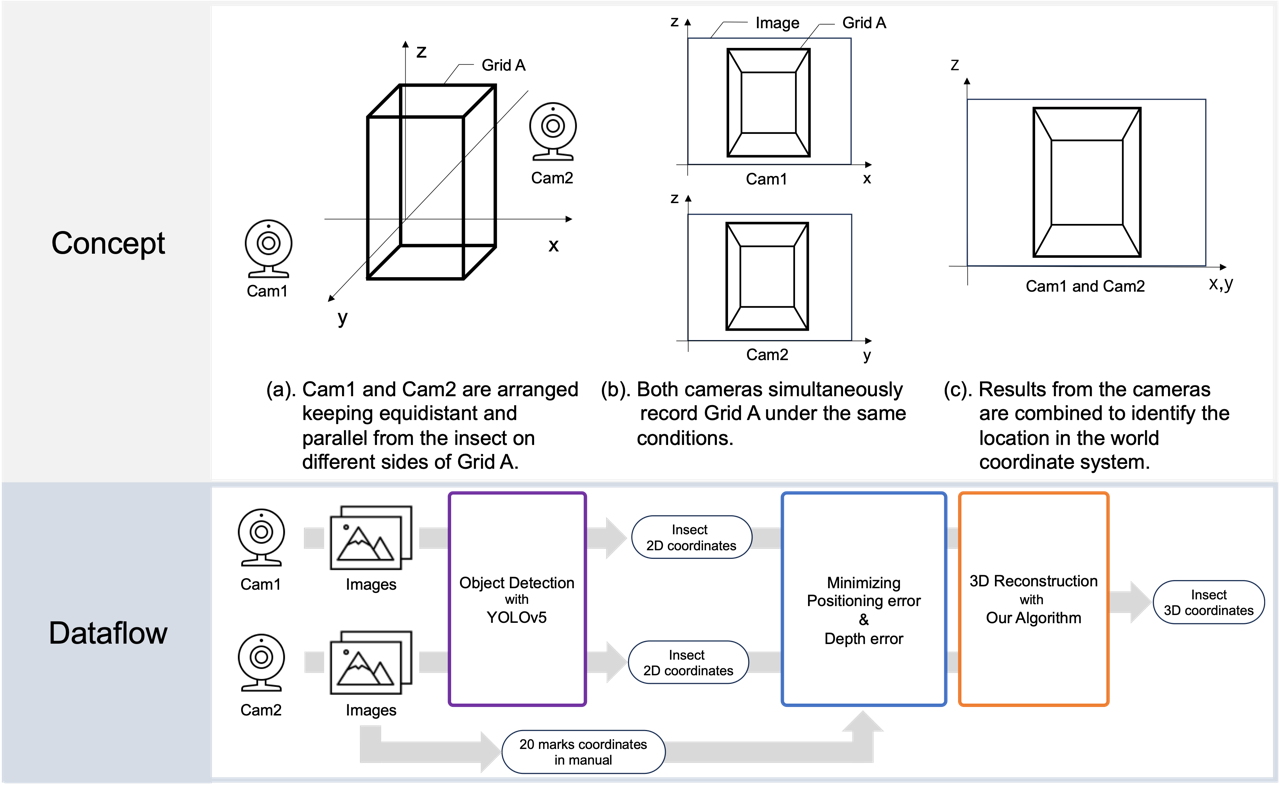}
			\caption{The concept of our algorithm and the dataflow.}
			\label{fig:summary}
			\vspace{-1.4em}
		\end{minipage}
	\end{figure*} 
		After constructing the dataset, we proceeded to train the YOLOv5 deep neural network specifically for the detection of individual the insects within each frame. During the training process, we utilized bounding box regression loss, object loss, and classification loss. The bounding box regression loss, calculated using the CIoU loss, measures the variance between the predicted bounding boxes and the ground truth bounding boxes for the insects within the image. The object loss, calculated with binary cross-entropy (BCE) loss, evaluates the model’s confidence in predicting the presence of the insect within the image. The classification loss, determined through BCE loss, measures the distinction between predicted class probabilities and the ground truth class probabilities. The total loss employed in training YOLOv5 constituted a weighted sum of the losses. \par
		For training the insect detector, we used a computer with an Intel(R) Core(TM) i9-10900K 10-core processor, RAM 64 GB, and the graphic card NVIDIA GeForce RTX 3090. We trained the model through 333 complete iterations, known as epochs, using a learning rate of 0.01. The learning rate gradually decreased utilizing the linear LR scheduler. We evaluated precision, recall and mAP as the indicator throughout the training process. The calculations of precision and recall as follows: 
	\begin{dmath}
		\text{Precision}=\frac{\text{True Positive}}{\text{True Positive} + \text{False Positive}}
	\end{dmath}
	\begin{dmath}
		\text{Recall}=\frac{\text{True Positive}}{\text{True Positive} + \text{False Negative}}
	\end{dmath}
		where a True Positive (TP) represents the total count of accurate positive predictions, meaning that instances where the model correctly identifies a pixel labeled as part of an insect image and accurately places it within a bounding box. A False Positive (FP) denotes instances where the model predicts a pixel to be within the bounding box, but it does not belong there. A False Negative (FN) represents the total count of predictions where the model indicates a pixel is outside any bounding box, but it actually occurs inside one. \par
		To prevent from overfitting the dataset, we set early stopping that stops the training if the accuracy of inference is constant until 30 epochs during the training. The accuracy indicator (fitness) is defined below: 
	\begin{dmath}
		\text{fitness}=0\times{\text{Precision}}+0\times{\text{Recall}}+0.1\times{\text{mAP@.5}}+0.9\times{\text{mAP@.5:95}}
	\end{dmath}
		where mAP@.5 shows a mAP at an Intersection over Union (IoU) threshold of 0.5 and mAP@.5:95 means a mAP at IoU threshold from 0.5 to 0.95, which is a more precise indicator than the former because the bounding box surrounds more than half of the insect’s. Finally, the YOLOv5 model training was completed in 303 epochs. After training, we achieved results of 0.93, 0.89, 0.91 and 0.38 corresponding to precision, recall, mAP@.5 and mAP@.5:95, respectively.
	\subsubsection{Insect Location Identification in Images}
		After the insect detector was trained using the collected dataset following the steps described above, we proceeded to input each of the frames taken from the video into the insect detector, obtaining the 2D location of the insect within the image. Then, the location of the insect at each particular frame as identified by the timestamp was saved into a CSV file. In this way, we were able to create corresponding location data for each of the selected frames.
	\subsection{3D Insect Spatial Reconstruction using 2D coordinates}
	\label{3DPlottingAlgorithm}
	\begin{figure*}[htbp]
		\begin{minipage}{\linewidth}
			\centering
			\includegraphics[width=0.7\columnwidth]{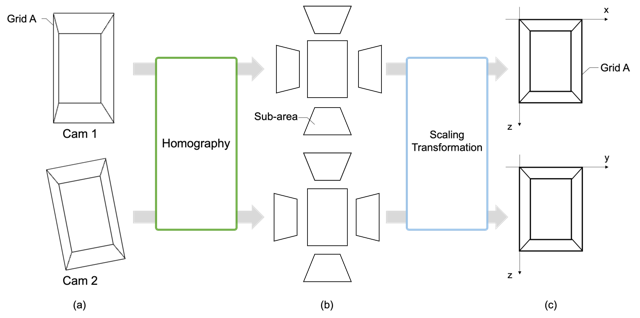}
			\vspace{-0.8em}
			\caption{Process of minimizing positioning error. (a) The images of Grid A are not same due to the difference of camera’s angle and position. (b) The images are adjusted for the shape and size of each sub-area. (c) The sub-areas are combined to be one rectangle. These rectangles of each of the images have same size and shape.}
			\label{fig:loacatemod}
			\vspace{-1em}
		\end{minipage}
	\end{figure*} 
		After the trained YOLOv5 model detected insects and saved their location on the CSV file, we constructed their 3D location with our plotting algorithm. A summary of the plotting algorithm is shown in Figure~\ref{fig:summary}. The coordinates on the CSV file use an image coordinate system that corresponds to a pixel in the displayed image. These coordinates are recalculated to transform the value to one on the world coordinate system using relative distance from insect to grid corner as an origin when the cameras surrounding Grid A are equidistant and straight on respective to each of Grid A’s sides. In Figure~\ref{fig:summary}, for example, coordinates on the \textit{xz} axis recorded by Camera 1 (Cam1) and coordinates on \textit{yz} axis recorded by Camera 2 (Cam2) refer to corresponding points on the same world coordinate system if both cameras are equidistant from their respective side of Grid A and the camera lens is directly aligned and straight on. Finally, the 3D location can be computed when both of the following requirements are satisfied: (1) two of the four side cameras detect the insect, (2) these cameras are located at a 90\textdegree~angle to each other, or on adjacent sides of Grid A; the combination of two of the four side cameras can be varied as long as they are not directly opposite each other. \par
		While the insect 3D location can be obtained following the above conditions, there are factors that cause some errors to occur. The following two types are representative of errors which most greatly affect the plotting algorithm: positioning error and depth error. (1) Positioning error occurs when the camera is placed slightly out of position and not straight on. Figure~\ref{fig:loacatemod}(a) depicts two snapshots taken by two of the four side cameras. Although the lower snapshot is more visibly misaligned, both snapshots will inevitably contain some degree of error. (2) Depth error has a negative effect on our measurements in that the greater the distance between the camera and insect, the more the insect appears closer to the center of a two-dimensional plane. The effects of this error can be minimized using the top camera, details of which are described in section~\ref{DepthMod}. 
	\subsubsection{Minimizing Positioning Error}
\begin{figure}[htbp]
		\centering
		\includegraphics[width=1.0\columnwidth]{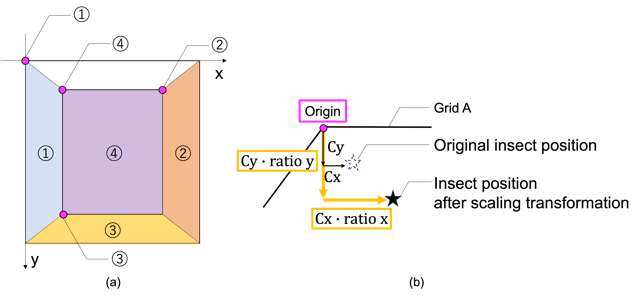}
		\vspace{-1.8em}
		\caption{Image of one side of Grid A and the sub-areas that undergo homography and scaling transformation to minimize positioning error. The origin for each respective sub-area is numbered accordingly.}
		\label{fig:persearea}
		\vspace{-1.4em}
	\end{figure}

	\paragraph{Homography}
		To minimize positioning error, the four side cameras are unified using homography. Homography is a two-dimensional projective transformation that maps points from one plane to another. The homography matrix H transforms 2D points as follows:
	\begin{align}
		\left( \begin{array}{r}
			x'_{1}\\ x'_{2} \\ x'_{3}
		\end{array} \right) = \begin{bmatrix}
			h_{11} & h_{12} & h_{13} \\ h_{21} & h_{22} & h_{23} \\ h_{31} & h_{32} & h_{33}
		\end{bmatrix} \left( \begin{array}{r}
			x_1 \\ x_2\\x_3
		\end{array} \right),\; \therefore x' = Hx
	\end{align}
		where $h_{ij}$ are parameters acquired in the process of the camera’s calibration, $x_k$ are coordinates on image coordinate system before processing, and $x'_l$ are coordinates after processing. The goal is to unify the views of the four side cameras as shown Figure~\ref{fig:loacatemod}(c). Each of the two grids looks like a larger rectangle consisting of five sub-areas: four trapezoids surrounding a smaller rectangle in the recorded videos. The coordinates of the four corners of each of the sub-areas are obtained manually using white stickers placed on the grid as markers. Using these coordinates, homography was applied to each of the sub-areas (Figure~\ref{fig:persearea}(a)). These sub-areas are adjusted in shape and size (Figure~\ref{fig:loacatemod}(b)), and the calculation is saved as a matrix in NPY file format (NumPy version is 1.22.1). After acquiring the matrix for each of the four sub-areas, we multiplied the matrix and the insect’s coordinates. 
	\paragraph{Scaling Transformation}
		Although all sub-areas are redefined by homography, we have not yet reached the stage depicted in Figure~\ref{fig:persearea}(c) due to mismatch in length of sides of the sub-areas to be connected. To combine these sub-areas, we applied scaling transformation after homography. In this second transformation, the ratio of scaling (ratio x and ratio y) is calculated using information derived from the white markers and camera pixel size, which is saved in TXT format (Figure~\ref{fig:persearea}(b)). After acquiring the ratio of scaling for each of the four sub-areas, we multiplied the ratio and the distance between the insect’s coordinates. 
	\paragraph{Re-mapping to the World Coordinate System}
		After undergoing homography and scaling transformation, an origin to be used as a point of reference for identifying the location of the insect in each sub-area was determined. Next, location data of the insect in the sub-area is transferred to a consolidated map of the four sub-areas that contains one common origin, which is referred to as the Model Grid (MG). When this mapping process is done with data from any two of the four side cameras that are located on adjacent sides of Grid A, a single 3D location point can be determined for mapping the insect on the world coordinate system.
	\subsubsection{Depth Error Modification}
	\label{DepthMod}
		Depth error has two properties. We considered the following correction methods for each.\par
\begin{figure}[htbp]
		\vspace{-1.5em}
		\centering
		\includegraphics[width=1.0\columnwidth]{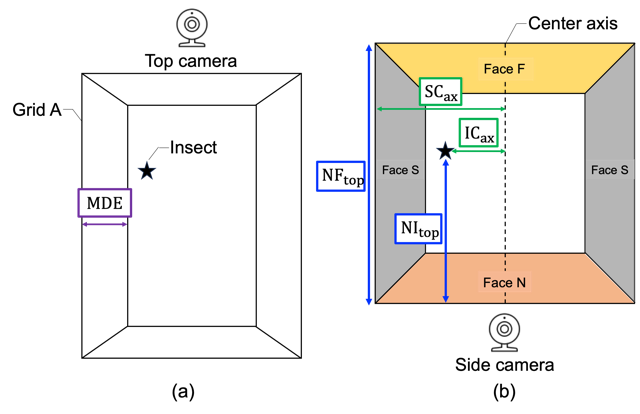}
		\vspace{-1.0em}
		\caption{How the adjustment for depth error is calculated.  (a) Side view. (b) Top view.}
		\label{fig:depthmod}
		\vspace{-2.0em}
	\end{figure}

		The first property is related to parallax where objects appear to move closer to the center point of the frame as their distance increases from the camera or foreground. For example, in the still frames of Grid A taken by the side cameras, the plane or face nearest to the camera (denoted as Face N) appears larger in size while the face farthest from the camera (denoted as Face F) appears relatively smaller, even though both faces actually have equal shape and size, and are positioned linearly. This perspective illusion causes what erroneously appears to be a shift in the location of the insect towards the center of the frame, although its actual change in movement is only in moving away from the camera, with no change in lateral or vertical direction. This error decreases as the insect on Face F approaches Face N, reaching a value of zero at Face N. To correct for this apparent shift or depth error, we measured the maximum distance of the shift between Face N and Face F as seen on the still frame image taken with the side camera. This is the Maximum value of the Depth Error, denoted here as MDE (Figure~\ref{fig:depthmod}(a)). We then took measurements from top camera images, using Face N as the basis for measurements, and determined the shortest distance from Face N to Insect ($\mathrm{NI_{top}}$ distance, Figure~\ref{fig:depthmod}(b)), divided by total distance from Face N to Face F ($\mathrm{NF_{top}}$ distance, Figure~\ref{fig:depthmod}(b)), and then multiplied by the MDE as measured from the side camera images to calculate the degree of error or Depth Error Factor (DEF), which is represented by the following formula:
	\begin{dmath}
		\text{DEF}=\text{MDE} \times \frac{\mathrm{NI_{top}}}{\mathrm{NF_{top}}}
	\end{dmath}
		This DEF was then used to correct the measured coordinate of the insect, details of which are described next. \par
		The second property of depth error is that degree of error decreases as the object is located closer to the axis running from Face N to Face F in the center of Grid A taken by the side camera, with an error value of zero when the object is located directly on the aforesaid center axis. Therefore, the DEF obtained in the First Property described above is at the maximum value possible when the object is at the farthest point from the center axis on either of the remaining two dimensions within the range of Grid A in a still image taken with the side camera. As the insect approaches the center axis, DEF decreases, and when the insect is at the center axis of the image, DEF becomes zero. Considering this second property, we used two measurements taken from top camera images: (1) the shortest distance between the insect and the center axis ($\mathrm{IC_{ax}}$), and (2) the distance from the center axis to either one of its parallel sides, or Face S ($\mathrm{SC_{ax}}$) as shown Figure~\ref{fig:depthmod}(b). We then divided $\mathrm{IC_{ax}}$ by $\mathrm{SC_{ax}}$ and multiplied the quotient by the DEF to obtain the final depth error adjustment value which was applied to the insect’s location coordinate as measured from the side camera still frames.
	\begin{dmath}
		\text{Final Depth Error Adjustment Value} = \text{DEF} \times \frac{\mathrm{IC_{ax}}}{\mathrm{SC_{ax}}}
	\end{dmath}
	\section{Result}
	\vspace{-0.5em}
	\subsection{Insect Detection via YOLOv5 Model}
\begin{figure}[htbp]
		\centering
		\vspace{-0.5em}
		\includegraphics[width=0.8\columnwidth]{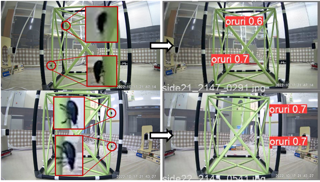}
			\vspace{-0.5em}
		\caption{Detection results using YOLOv5 (left: label, right: inference by trained YOLOv5 model).}
		\label{fig:YOLOresult}
		\vspace{-1em}
	\end{figure}

		We input the 548 test frames into the trained YOLOv5 model to predict the bounding boxes for the insects within each test frame. Figure~\ref{fig:YOLOresult} shows the detected bounding box in the test run after training. In the test data, scores of 0.94 and 0.39 for mAP@.5 and mAP@.5:95 were achieved respectively, with attained precision at 0.92 and recall reaching 0.92. The result of a confusion matrix is shown in Table~\ref{tab:ConfMtx}. 
	\begin{table}[htbp]
		\centering
		\vspace{-1.0em}
		\caption{Confusion matrix.}
		\vspace{-1.0em}
		\includegraphics[width=0.7\columnwidth]{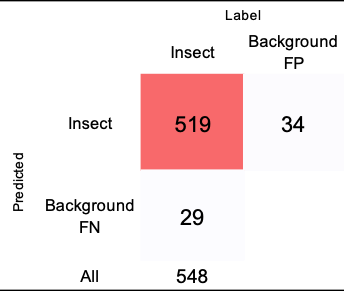}
		\label{tab:ConfMtx}
		\vspace{-2.0em}
	\end{table}
	\subsection{Visualizing Insects in 3D Space}
\begin{figure}[htbp]
		\centering
		\includegraphics[width=0.55\columnwidth]{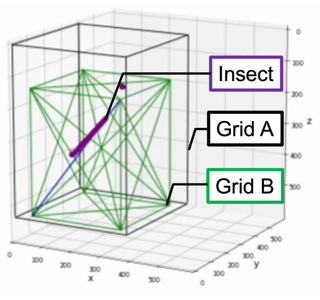}
		\vspace{-0.5em}
		\caption{3D visualization of the insect location.}
		\label{fig:visualize}
		\vspace{-0.8em}
	\end{figure}

		For each frame of the video, the trained YOLOv5 model detected the insect and saved the coordinates on a CSV file. The result of 3D insect visualization using information on the CSV files is shown in Figure~\ref{fig:visualize}. We used Matplotlib (version 3.5.3) for drawing three-dimensional plots. 
	\subsection{Evaluation of the 3D-Plotting Algorithm}
		\begin{figure}[htbp]
			\centering
			\vspace{-1em}
			\includegraphics[width=1.0\columnwidth]{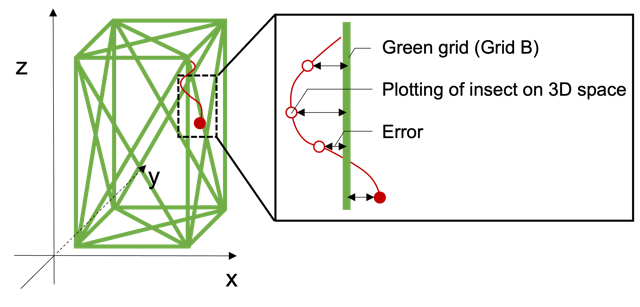}
			\vspace{-2em}
			\caption{Image of algorithm test. Distance between the 3D coordinates of insects and the surface of Grid B indicates degree of plotting error.}
			\label{fig:testmethod}
			\vspace{0em}
		\end{figure}
\begin{figure}[htbp]
		\centering
		\vspace{-1.5em}
		\includegraphics[width=1.0\columnwidth]{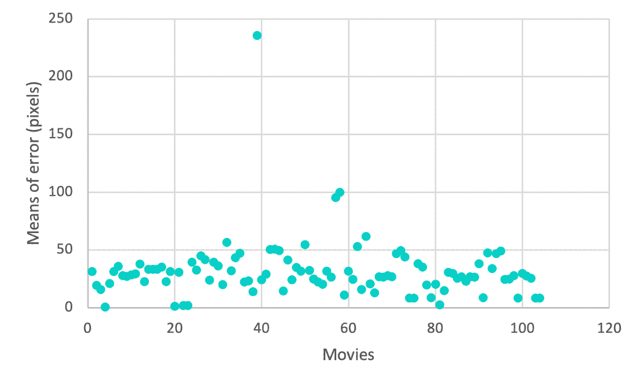}
		\vspace{-1.8em}
		\caption{Algorithm test result.}
		\label{fig:testresult}
		\vspace{-1.8em}
	\end{figure}

		We tested the plotting algorithm to evaluate how precisely it plots the insect’s location in virtual 3D space through the following the procedure. The insect walks on the frame of Grid B. Here, we constructed Grid B in three-dimensional space. If the insect’s location on Grid B could be plotted showing it in direct contact with the surface of Grid B, we would have evidence that the plotting algorithm was accurate. We defined the algorithm's plotting error as the distance between the 3D coordinates of insects and the surface of Grid B (Figure~\ref{fig:testmethod}). Therefore, we divided the test videos into segments based on the different surfaces of Grid B that insects walked on. For each segment, we calculated the average plotting error of the plotting algorithm for each video segment. Finally, we obtained the overall average from all videos and defined this as the plotting algorithm’s accuracy indicator. The results of the test are shown in Figure~\ref{fig:testresult}. Of the instances that the YOLOv5 was able to detect the location of insects in the side camera still frames, the percentage of insects that the plotting algorithm could plot in three-dimensional space was calculated to be 67\%. Furthermore, the accuracy of the plotting algorithm was 25.94 pixels, which corresponds to 2.594 centimeters in the real world.
	\section{Discussion}
	\vspace{-0.5em}
	\subsection{Summary and Comparison with Previous Works}
		In this study, we have presented an automated plotting algorithm for insect 3D tracking. The plotting algorithm is widely applicable for generating standardized monitoring data in limited experimental environments. It is useful for insect monitoring in experimental environment for four reasons. First, a video of an insect is analyzed automatically after recording, which reduces labor-intensive work for investigating insect behavior. Second, the plotting algorithm can be used for a variety of environments because of the ability of YOLOv5 to learn generalized representations of objects. Some studies in 3D tracking of insects fix or adjust background environments to extract small insects using differences in pixel values. The plotting algorithm is not as limited by background factors because the YOLO model is trained not only in the features of insects but also in the background information~\cite{Redmon2015}. Third, incorporating our plotting algorithm enhances the accuracy of insect monitoring by integrating depth information. In contrast, conventional 2D-tracking algorithms exclusively track movement along the x-horizontal and y-vertical axes, disregarding the z-depth dimension. This limitation becomes evident when an insect moves directly towards or away from the camera. The 3D-plotting algorithm can precisely detect motion in the depth dimension by minimizing positioning and depth error, and thereby offering invaluable insights, especially in scenarios where motion along the z-direction holds significance. Fourth, the 3D-plotting algorithm is available with inexpensive hardware. We used a commercially-available camera as a recording device, which makes automatic insect observation more accessible and affordable. However, we must note that obtaining the necessary computational hardware for training the YOLOv5 can be somewhat costly due to its usage of deep learning technology. 
	\subsection{Limitation}
	\subsubsection{Detection Accuracy}
		Our plotting algorithm ultimately depends on YOLOv5’s detection accuracy. Although all cameras can record an insect’s activity, all recorded insect’s coordinates might not be shown in virtual 3D space for two reasons: a lack of training on the YOLOv5 model and insect occlusion. The YOLOv5 model cannot detect insects when it is not trained using datasets of adequate quality and quantity. It is recommended that datasets for training be prepared following the tips for best training results on YOLOv5’s training manual~\cite{Jocher2021}. Further, satisfactory results cannot be obtained even if the YOLOv5 has high detection accuracy because of an insect occlusion. We recommend putting no obstacles in the grid to record insects clearly. On the other hand, it is possible that some of these problems may be resolved to some extent in the future with improvements in the prediction accuracy of object detection AI and advances in learning algorithms. 
	\subsubsection{Effect of Insect Density}
		Our plotting algorithm can be used to monitor behavior of single insects, but the simultaneous occurrence of two or more insects decreases the accuracy considerably. The plotting algorithm currently does not have the ability to distinguish several insects recorded at the same time. However, increasing the number of cameras, modifying the 3D algorithm, or implementing multi-object tracking technology may contribute to a partial solution to this problem. 
	\subsubsection{Sensitivity of Equipment's Arrangement during Recording}
		One of the more challenging aspects using our plotting algorithm is sensitivity of cameras’ arrangement during recording. Although it is stipulated that the position of the cameras be at the same height, distance, and angle from the grid, slight misarrangement negatively affects the accuracy of 2D coordinates, which are calculated from the recorded images and used in the 3D plotting results. We recommend anchoring the cameras and other recording equipment for the recording process. If this equipment is accidentally moved during recording, it is necessary to get the equipment properly aligned again. One possible future solution could be to build a dedicated, fixed experimental environment and increasing the number of cameras, or using self-localization algorithms. 
	\section{Conclusion}
		In this study, we developed a precise, automatic 3D-insect tracking algorithm using five commercially available and affordable cameras and the YOLOv5 object detection model. We proposed a method to correct errors in location measurement and plotting caused by camera placement and parallax, enhancing the accuracy of recording insect position. We quantitatively assessed the algorithm's performance in virtual 3D space, providing objective evidence of its practicality. The utilization of this system is expected to contribute to ongoing ecological studies by enabling researchers to automatically track and analyze patterns of insect behavior and movement, thus, elucidating the crucial relationship between insects and their environment.
		
	\subsection*{Acknowledgement}
	The authors would like thank Margaret Chang for her invaluable contribution and support in the writing of this English manuscript.
	\subsection*{CRediT Authorship Contribution Statement}
		\noindent
		\textbf{Daisuke Mori}: Conceptualization, Data curation, Investigation, Methodology, Software, Visualization, Writing – original draft.
		\textbf{Hiroki Hayami}: Investigation, Resources.
		\textbf{Yasufumi Fujimoto}: Investigation, Resources.
		\textbf{Isao Goto}: Conceptualization, Funding acquisition, Investigation, Methodology, Project administration, Resources, Supervision, Writing – review \& editing.
	\bibliographystyle{unsrt}
	\bibliography{main}
\end{document}